\documentclass[
]{ceurart}

\usepackage{listings}
\usepackage{ulem}
\usepackage{soul}
\usepackage{xcolor}

\begin{document}

\copyrightyear{2026}
\copyrightclause{Copyright for this paper by its authors.
  Use permitted under Creative Commons License Attribution 4.0
  International (CC BY 4.0).}

\conference{Ital-IA 2026: 6th National Conference on Artificial Intelligence, organized by CINI, June 18-19, 2026, Rome, Italy}

\title{A Formal Framework for Declarative Agentic AI in Business Process Analysis}

\author[1]{Mohammad Azarijafari}[%
orcid=0000-0002-3018-7000,
email=mohammad.azarijafari@unitn.it,
]
\cormark[1]
\fnmark[1]
\address[1]{Department of Industrial Engineering, University of Trento, Via Sommarive 9, 38123, Trento, Italy}

\author[1]{Luisa Mich}[%
orcid=0000-0002-0018-6883,
email=luisa.mich@unitn.it,
]
\fnmark[1]

\author[2]{Michele Missikoff}[%
orcid=0000-0002-7972-5201,
email=michele.missikoff@iasi.cnr.it,
]
\fnmark[1]
\address[2]{Istituto di Analisi dei Sistemi ed Informatica (IASI) “Antonio Ruberti”, National Research Council (CNR), Via dei Taurini 19, 00185, Rome, Italy}

\cortext[1]{Corresponding author.}
\fntext[1]{These authors contributed equally.}

\begin{abstract}
  Agentic AI opens new opportunities for automating Business Process (BP), enabling autonomous decision-making and dynamic adaptation. However, realising this potential requires BP entities and their interactions to be defined with formal precision. This paper presents a formal framework for Agentic BP analysis through the AGO methodology. AGO captures the modelling perspective in terms of who is acting (Agents), why it is carried out (Goals), and what the relevant entities are (Objects). Grounded in set theory and mathematical logic, we formally define the AGO entity types and their interactions, organising all definitions into a BP Knowledge Base (BPKB). The resulting BPKB supports structured querying, incremental updates, and automatic generation of BP workflows, while ensuring soundness and completeness of the derived paths.

\end{abstract}


\maketitle

\section{Introduction}

Business processes (BPs) are at the core of modern organisations, governing how work is coordinated, decisions are made, and value is delivered \cite{Weske}. As organisations grow in scale and complexity, facing increasing pressure to adapt, automate, and remain competitive, the need for rigorous, systematic, and formally grounded methods for BP analysis becomes increasingly critical \cite{Schuler,Ahmad}. The recent rise of agentic AI \cite{acharya2025agentic} introduces new opportunities for automating and enhancing BP execution. However, realising this potential requires that BP entities and their interactions be defined with formal precision, so that agentic components can operate on a well-structured and verifiable knowledge base.

Traditional methods for BP analysis are based on a diagrammatic approach \cite{Benedict,Dumas}, in which the analysis primarily consists of identifying tasks and connecting them to form a directed graph that represents the BP workflow. Widely adopted notations, such as BPMN diagrams, exemplify this paradigm. While broadly used in both academia and industry, such methods present well-known limitations: they require considerable technical expertise to construct and interpret, are difficult to decentralise across business units, produce artefacts that are hard to query, maintain, or evolve, and hinder the direct involvement of non-technical business stakeholders.

To address these limitations, the AGO (Agents, Goals, Objects) methodology, was introduced in \cite{ital-ia25,agentics25}, proposing a declarative approach, based on conceptual modelling of business entities, and knowledge representation. The AGO analysis starts from the identification and modelling of three basic business entities: objects, goals, and agents. They serve as building blocks to reconstruct, bottom-up, the full workflow of a BP, deriving all its paths from start to end without the need to build the workflow in its entirety. The central artefact of AGO analysis is the Business Process Knowledge Base (BPKB), which stores all identified entities, their relationships, together with consistency constraints.

The AGO methodology presents several advantages. Its decentralised analysis allows different business units to collect local knowledge independently, ensuring a virtuous separation of concerns. The methodology's simplicity also enables business people without technical skills to participate directly in conceptual modeling and BP design, reducing their dependence on technical staff and improving business-IT alignment. Storing all BP knowledge in declarative form into the BPKB further supports querying, collaborative analysis, and easing corrections, updates, and evolutions. However, a rigorous formalisation of AGO remains a key requirement to fully realise these advantages.

In this paper, we propose a formal framework grounded in set theory and mathematical logic \cite{Preparata,Barwise} for BP analysis, providing rigorous support for AGO to enable agentic AI applications. We firstly identify the three core entity types, objects, goals, and agents, and organise them into a Lexicon and Glossary as the first two steps of the analysis. Next, We introduce a formal characterisation of goal types, distinguishing conjunctive, inclusive disjunctive, and exclusive disjunctive goals, each with a precise satisfaction condition. Then, we define agents through their trigger and delivery sets. Finally, in the third step, all the collected knowledge is stored in the resulting BPKB, grounded in formal theory. This enables automatic generation of the whole workflow while verifying its soundness and completeness.

The rest of the paper is organised as follows. Section \ref{sec:entities} presents the three core AGO entities. Section \ref{sec:bpkb} describes the resulting BPKB and the derivation of the BP workflow. Section \ref{sec:conclusion} concludes the paper with directions for future work.

\section{The AGO Methodology}
\label{sec:entities}

The AGO methodology represents an important shift in the BP analysis, moving the focus from how a process is executed to (i) \textit{what} the relevant business entities are, (ii) \textit{what} the process intends to achieve, and (iii) \textit{who} is in charge of the operations. This ontological shift is supported by a consistent conceptual framework that distinguishes between active and passive entities, and a formal framework that provides rigorous definitions of the BP entities. This paper extends the first version of AGO \cite{agentics25}.

Thus, in the AGO, business reality is represented by a set of \textit{objects} that model the evolving state of affairs, a set of \textit{goals} representing states of affairs to be reached during process execution, and a set of \textit{agents} representing the active components of the BP operating to reach the goals.

According to AGO, the analysis proceeds following three steps, aimed at producing structures with an increasing knowledge content. The first two steps tackle the initial structures, Lexicon and Glossary, that are highly intuitive to ease the involvement of business people. The third step tackles the Business Process Knowledge Base, aimed at organising and storing the BP knowledge, by including and expanding the content of the two previous structures. Below, the first two steps are formally presented, adopting the simple pizza shop example introduced in \cite{agentics25}, while the third step is addressed in Section \ref{sec:bpkb}.

\vspace{-0.2cm}
\paragraph{Step 1: Lexicon.}

The first structure is a Lexicon that collects all the terms denoting the relevant business entities. The terminology is organised according to the AGO categories. With reference to a pizza shop example, we have the following entries in the lexicon:

\vspace{0.2cm}
\smallskip
\noindent \uline{Objects ($O$)}: \textit{ClientOrder, AcquiredOrder, InvalidOrder, ValidOrder, OrderedPizza, CreditCardData, Payment, ReadyPizza, RejectedOrder, CookAlert, CliHomeData, FulfilledOrder}

\smallskip
\noindent \uline{Goals ($G$)}: \textit{OrderSubmitted, OrderAcquired, OrderRejected, CookInformed, PizzaReady, OrderFulfilled}

\smallskip
\noindent \uline{Agents ($A$)}: \textit{Client, ShopAdmin, Cook, DeliveryBoy}

\vspace{0.4cm}
\noindent In formal terms, we have:
\vspace{-0.3cm}
\paragraph{Definition 1: Lexicon.}

Given a finite set of business terms $N$, a Lexicon L is a triple $L := (O, G, A)$ where $O$, $G$, and $A$ contain the terms denoting business objects, goals, and agents, respectively. They represent three mutually disjoint partitions of $N$.

\vspace{-0.3cm}
\paragraph{Step 2: Glossary.}

In the second step, we build the glossary $G$, where each term in $L$ is associated with a label that represents a concept identifier, and a description. The Glossary is organised in three segments, $OG$, $GG$, $AG$, corresponding to the three partitions of $L$, each of which has a cardinality equal to the corresponding Lexicon cardinality.

\vspace{-0.3cm}
\paragraph{Definition 2: Glossary.}

Starting from the terms $t \in L$, the Glossary $Glo$ is a triple defined as follows:
\[
\mathit{Glo} := (OG,\, GG,\, AG)
\]
where each segment contains object, goal, and agent entries, respectively.

Each entry has the following structure: $(EID, Na, La, De)$, where $EID$, stands for a unique type identifier $OID$, or $GID$, or $AID$. $Na$ is a term $t \in L$ that assumes the role of entity name, $La$ and $De$ are the entity label and description, respectively. 

\noindent For instance, in our example:\\

\vspace{-0.2cm}
$
(O_1, Na: ClientOrder, La: CliOrd, De: \textit{``Order issued by a client''})
$

$
(G_1, Na: OrderAcquired, La: OrdAcq, De: \textit{``The order has been acquired''})
$

$
(A_1, Na: Client, La: Cli, De: \textit{``The person who submits an order''})
$

\vspace{0.3cm}
The notions of objects, goals, and agents play a strategic role in BP specification and fundamentally shape the operational rationale of AGO. Before proceeding to the construction of the BPKB, each of these components is elaborated in the following subsections.

\subsection{Business Objects}

Business objects are the passive entities of a BP, representing the relevant aspects of the business reality.

\vspace{-0.3cm}
\paragraph{Definition 3: Business object.}

In the analysis phase we firstly identify the types of the business objects, $O:= {O_1, …, O_n}$, that represent the business reality in the various configurations assumed during the BP execution. In AGO, objects are the elements used to define the goals. During a BP execution, agents, aiming at satisfying goals, generate new objects that represent the evolving reality. Therefore, a BP execution is reflected by the business goals that are achieved in a sequence.

In the analysis phase, business objects are described at the Glossary level. Their full definition, including structure, attributes, and relationships, will be provided later in the design phase.

\subsection{Business Goals}
\label{sec:goals}

Goals are strategic entities with a double nature. From a business perspective, they represent sought-after states of affairs. From the operational perspective, they are necessary for the cooperation and coordination of agents. Furthermore, goals’ contents are used by agents to exchange information.

A goal is specified by a set expression, defined over a subset of the business objects $O$, that states a satisfaction condition. In particular, we have three basic sorts of goal expression: \textit{conjunction}, \textit{inclusive disjunction}, and \textit{exclusive disjunction}. Intuitively, at runtime, a conjunctive goal is achieved if all of its declared object types are instantiated. In disjunctive goals, the object set is organised as a cover\footnote{A cover of X is a collection of subsets $\{C_1, C_2, …\}$ of $X$ whose union equals $X$.}. For inclusive disjunction, the goal is achieved if one or more cover elements are instantiated. For an exclusive disjunctive goal, its achievement requires that one and only one covering subset is instantiated.

For instance (simplifying the notation), having goals $G_1$, $G_2$, $G_3$ of the three sorts, $G_1 = \{O_1, O_2\}$ is conjunctive; $G_2 = \{O_3 \text{ or } O_4\}$, is inclusive disjunctive, and $G_3 = \{O_3 \text{ xor } O_4\}$ is exclusive disjunctive. In more precise terms, we have the following definitions.\\

\vspace{-0.3cm}
\noindent\textbf{Definition 4: Conjunctive business goal.}
A conjunctive business goal $G_c$ is defined by a set of object types and a satisfaction condition. Thus, we introduce the Boolean functions: $\mathit{Sat}(G)$. It is evaluated at runtime and, checking if all declared object types are correctly instantiated, determines whether the goal is achieved.

Given a goal $G_c := \{O_1, O_2, \ldots, O_j\}$, it is satisfied by a goal instance $g_c$ if and only if the following expression holds.
\[
\mathit{Sat}(G_c) \;\Leftrightarrow\; g_c \in G_c,\; g_c = \{o_1, \ldots, o_j\} \;\text{and}\; o_1 \in O_1, \ldots, o_j \in O_j.
\]

\vspace{-0.2cm}
\paragraph{Definition 5: Inclusive Disjunctive business goal.}

An inclusive disjunctive goal $G_i$ is defined as a covering of a set of business objects, where the object types of at least one cover subset is instantiated. We introduce the $\mathit{Cov}$ function on a goal type, defined by the analyst. For instance, given a goal type $G_i = \{O_1, O_2, O_3, O_4, O_5\}$:
\[
\mathit{Cov}(Gi) = \{C_1, C_2\}, \;\text{where } C_1 = \{O_1, O_2, O_3\},\; C_2 = \{O_3, O_4, O_5\}.
\]
$G_i$ will be satisfied if and only if at least one cover subset is instantiated:
\[
\mathit{Sat}(G_i) \;\Leftrightarrow\; \mathit{Sat}(C_1) \text{ or } \mathit{Sat}(C_2).
\]

\vspace{-0.3cm}
\paragraph{Definition 6: Exclusive Disjunctive business goal.}

Given a goal $G_e = \{O_1, O_2, O_3, O_4, O_5\}$, and two cover subsets $C_1$, $C_2$ as defined above, $G_e$ is an exclusive disjunctive goal if and only if just one cover subset is instantiated. Here is a simplified formalisation:
\[
\mathit{Cov}(G_e) = \{C_1, C_2\}, \;\text{where } C_1 = \{O_1, O_2, O_3\},\; C_2 = \{O_3, O_4, O_5\}.
\]
\[\text{Then: }
\mathit{Sat}(G_e) \;\Leftrightarrow\; \mathit{Sat}(C_1) \text{ xor } \mathit{Sat}(C_2).
\]

To complete the definition of a goal, we need to indicate which agents are associated with the goal. Then we declare: \textit{incoming agents (InAg),} acting for goal satisfaction, and \textit{outgoing agents (OutAg),} triggered when the goal is satisfied.

Here, the definition of a goal, as represented in the BPKB:
\[
G := (GID,\, GNa,\, GLa,\, GDe,\, SatDef,\, InAg,\, OutAg)
\]

After the goal type identifier, the first three components come from the Glossary. $\mathit{SatDef}$ defines the satisfaction formula. $\mathit{InAg}$ and $\mathit{OutAg}$ denote the incoming and outgoing agents, respectively.

Disjunctive goals are referred to as \textit{fork goals}, since they have more than one outgoing agent, thus representing a fork in the workflow diagram. Symmetrically, when a goal has more than one incoming agent, we have a \textit{merge goal}. Therefore, the goals determine the workflow topology. It is not provided during the analysis, it is reconstructed bottom-up, at the end of the analysis, starting from the AGO entities, as discussed subsequently.

\subsection{Business Agents and Skills}
\label{sec:agents}

Business agents are the active elements of AGO. They `observe' the business reality and, when the triggering conditions are met, start their operations. Each agent has one or more skills that determine what operations it is able to perform. In the declarative perspective of the AGO methodology, an agent with its skills is a binary relation between an input set of objects, referred to as the trigger set ($TS$), and an output set, referred to as the delivery set ($DS$).

\[
A = \{(TS, DS)\}, \;\text{where } A \subseteq \mathcal{P}(O) \times \mathcal{P}(O).
\]

$A$ is a finite set of pairs $(TS, DS)$, each of which corresponds to a skill. $\mathcal{P}(O)$ represents the power set of $O$.

From an operational perspective, $TS$ indicates the condition for the agent to start its operations, while the $DS$ is the produced result. Therefore, we consider agents' skills to be functions that produce a new set of business objects: $A_s(TS_s) = DS_s$, where the subscript indicates the specific skill. $TS$ and $DS$ are contained in the \textit{source} and \textit{destination goals} of an agent, respectively.

The full agent definition is as follows.
\[
A := (AID,\, ANa,\, ALa,\, ADe,\; \{(SID,\, SNa,\, SLa,\, SDe,\, G_{in},\, TS,\, G_{out},\, DS)\})
\]

Where $AID$ is the agent identifier, $ANa$, $ALa$, and $ADe$ are the name, label, and description, respectively. Then, there is the set of agents' skills, each of which carries its own identifier, name, label, and description, mirroring the agent structure. Then, the trigger and delivery sets that apply to skills, and the input goal ($G_{in}$) and the output goal ($G_{out}$) where the $TS$ and $DS$ are included. Note that the trigger and delivery sets are specifically associated to the agent's skills. The $G_{in}$ of a skill must match a goal whose $\mathit{OutAg}$ includes that agent, and the $G_{out}$ must match a goal whose $\mathit{InAg}$ includes it. This bidirectional consistency is a key integrity condition of the BPKB.

\section{The Resulting BPKB and Derived Workflow}
\label{sec:bpkb}

\paragraph{Step 3: Business Process Knowledge Base (BPKB).} The BPKB is the final structure produced by the analysis. Having defined all the AGO entity types that participate in the BP, we add the consistency rules. Among others, we enforce that all goals need to be reachable and satisfiable, that objects cannot exist outside of goals, as well as trigger sets and delivery sets that need to be included in goals.

As anticipated, in AGO, the workflow diagram is not provided by the analysts. Following the formalisation of the entities given in Section~\ref{sec:goals} and ~\ref{sec:agents}, it is possible to reconstruct the full workflow with all its end-to-end paths. From the agent definitions available in the BPKB, we can extract (abstracting the skills) all the triples of the form $(G, A, G')$. Then, we obtain a precedence relation $\prec$ on goals, where $G \prec G'$ holds if and only if there exists an agent $A$ with a skill such that $TS \subseteq G$ and $DS \subseteq G'$. By iteratively applying chaining, we obtain goal sequencing and the workflow. For instance, given:

\[
(G_1, A_x, G_2), (G_2, A_y, G_3)
\]
\[\text{We derive: }
G_1 \prec G_2 \prec G_3
\iff
\exists A_x, A_y :
TS_x \subseteq G_1
\land
DS_x \subseteq G_2
\land
TS_y \subseteq G_2
\land
DS_y \subseteq G_3
\]

\paragraph{AGO as a formal theory.} In a formal perspective, the AGO elements collected in the analysis represent the axioms of a formal theory. Thus, it is possible to automatically build all the end-to-end paths of the workflow that represent the theorems. Such a rigorous approach allows us to prove important formal properties, in particular \textit{soundness} and \textit{completeness}. The first ensures that all the derived end-to-end paths are valid, i.e., they correspond to correct BP instances, and the second that all valid end-to-end paths can be derived. A full treatment of this part falls outside the scope of this paper.

\begin{figure}[h!]
    \centering
    \includegraphics[width=0.9\textwidth]{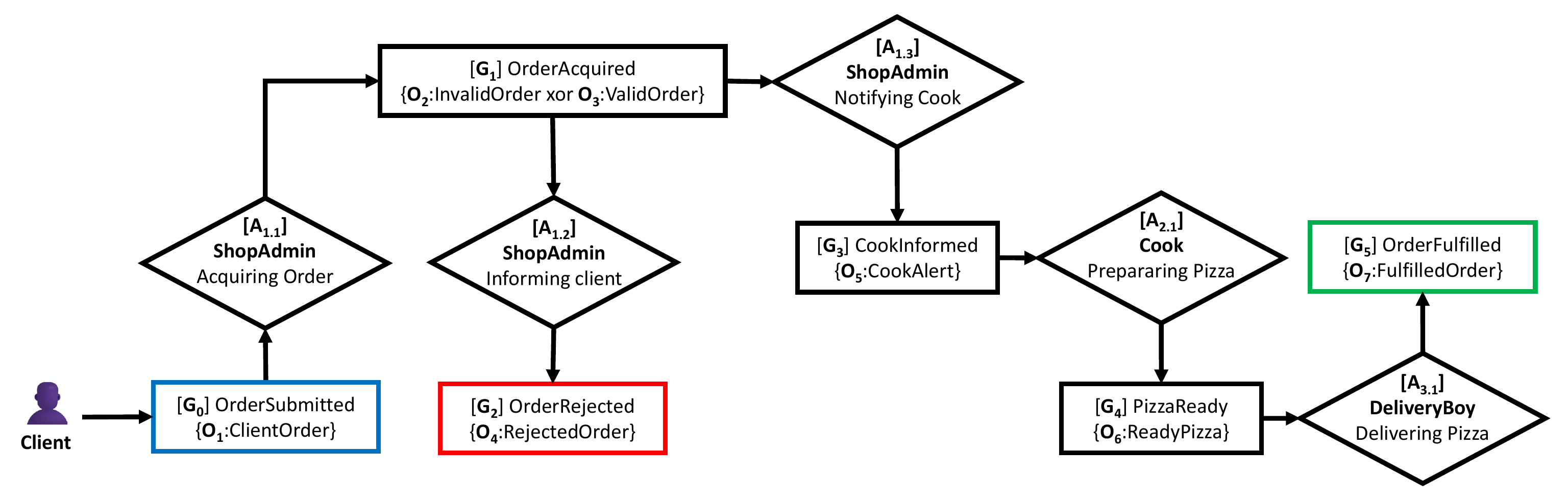}
    \caption{Agent-based workflow for the pizza delivery process.}
    \label{fig:BP diagram topology}
\end{figure}

The BPKB triples derived from the pizza shop example are presented below, where the asterisks denote the terminating goals. The corresponding workflow diagram is illustrated in Figure~\ref{fig:BP diagram topology}.

\[
(G_0, A_{1.1}, G_1), 
(G_1, A_{1.2}, G_2^*),
(G_1, A_{1.3}, G_3),
(G_3, A_{2.1}, G_4),
(G_4, A_{3.1}, G_5^*) 
\]

\section{Conclusion and Future Work}
\label{sec:conclusion}

In this paper, we introduced a formal framework for BP analysis based on AGO, a declarative agentic methodology comprising three types of entities: agent, goal, and object. This declarative paradigm not only reduces model complexity but also provides a formal foundation for agentic AI applications, providing enterprises with the flexibility and resilience needed to thrive in an increasingly dynamic world. The AGO unfolds across three phases of BP Developmennt (BPD): analysis, design, and implementation. In this paper, we focused on the analysis phase and formally defined a BPKB that stores a lexicon, a glossary of business terms, and the structures of AGO business entities. Grounded in set theory and mathematical logic, the BPKB enables the automatic generation of complex BP workflows.

Future work will address the two remaining phases of BPD. In the design phase, the BPKB is enriched with technical details, including class definitions and database schemas, yielding the BPKB+. In the implementation phase, agents' skills are provided with prompts that capture the application logic, which are fed to LLMs at runtime to achieve BP execution.

\begin{acknowledgments}
  This work has been partially supported by the PNRR MUR project PE0000013-FAIR CUP B53C22003630006, Next Generation EU program, Italian Government. 
\end{acknowledgments}

\section*{Declaration on Generative AI}
  During the preparation of this work, the authors used Claude Sonnet in order to: Grammar and spelling check. After using this tool, the authors reviewed and edited the content as needed and take full responsibility for the publication’s content. 

\bibliography{sample-ceur}

\end{document}